\pdfoutput=1

\documentclass[11pt]{article}

\usepackage[final]{coling}

\usepackage{times}
\usepackage{latexsym}

\usepackage[T1]{fontenc}

\usepackage[utf8]{inputenc}

\usepackage{microtype}

\usepackage{inconsolata}

\usepackage{graphicx}

\usepackage{booktabs}
\usepackage{multirow}
\usepackage{subcaption}
\usepackage{amsmath}

\newcommand{\q}[1]{``#1''} 

\title{Leveraging Explicit Reasoning for Inference Integration \\ in Commonsense-Augmented Dialogue Models}

\author{
  Sarah E. Finch \\
  Department of Computer Science \\
  Emory University \\
  Atlanta, GA, USA \\
  \texttt{sfillwo@emory.edu} \\
  \And
  Jinho D. Choi \\
  Department of Computer Science \\
  Emory University \\
  Atlanta, GA, USA \\
  \texttt{jinho.choi@emory.edu}
}

\begin{document}

\maketitle

\begin{abstract}

Open-domain dialogue systems need to grasp social commonsense to understand and respond effectively to human users. Commonsense-augmented dialogue models have been proposed that aim to infer commonsense knowledge from dialogue contexts in order to improve response quality. However, existing approaches to commonsense-augmented dialogue rely on implicit reasoning to integrate commonsense inferences during response generation. In this study, we explore the impact of explicit reasoning against implicit reasoning over commonsense for dialogue response generation. Our findings demonstrate that separating commonsense reasoning into explicit steps for generating, selecting, and integrating commonsense into responses leads to better dialogue interactions, improving naturalness, engagement, specificity, and overall quality. Subsequent analyses of these findings unveil insights into the effectiveness of various types of commonsense in generating responses and the particular response traits enhanced through explicit reasoning for commonsense integration. Our work advances research in open-domain dialogue by achieving a new state-of-the-art in commonsense-augmented response generation.

\end{abstract}

\section{Introduction}

\begin{figure}[t]
    \centering
    \includegraphics[width=\columnwidth]{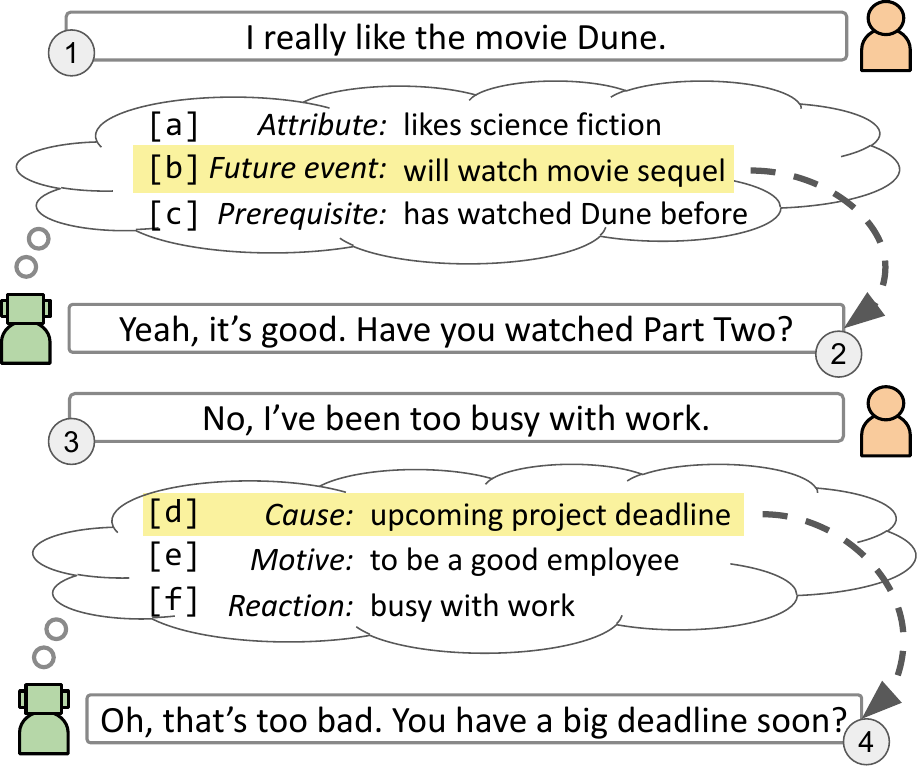}
    \caption{Example dialogue demonstrating the reasoning flow for integrating commonsense.}
    \label{fig:dialogue-commonsense-example}
    \vspace{-3ex}
\end{figure}

In open-domain dialogue, dialogue systems must engage in open-ended conversation with a human user, adapting fluently and intelligently to the topics that are introduced, which often involve discussions about life experiences \cite{robinson_museum_2008, fillwock:18, mitsuda:19}. 
As illustrated by the dialogue example in Figure \ref{fig:dialogue-commonsense-example}, meaningful follow-up responses are often driven by speculative thinking regarding the experiences shared by the human user \cite{finch_emora:20}, such as predictions about likely future plans of the user (\texttt{[b]}, Turn 2) and likely reasons behind the user's actions (\texttt{[d]}, Turn~4). This manner of inferential reasoning enriches the contextual understanding of the user's input and facilitates the generation of insightful responses. The ability to draw such inferences stems from shared worldviews and mutual experiences of humans—a phenomenon commonly known as \q{commonsense} \cite{clark:91}.

As such, the goal of commonsense-augmented dialogue modeling is to leverage useful commonsense inferences for producing more compelling and human-like responses in human-computer dialogue. For a given dialogue situation, there are numerous commonsense inferences that can be drawn since inferential commonsense has a many-to-many mapping due to its speculative nature \cite{shen:22,finch_convosense_2024}. Consequently, commonsense-augmented dialogue modeling is a reasoning process of (1) deriving commonsense knowledge that likely holds true for a given dialogue context, (2) identifying a subset of true commonsense that is appropriate for generating a response, and (3) synthesizing a response from the identified commonsense knowledge.
No previous work tackles all three of these components of commonsense-augmented dialogue modeling in an explicit manner. Often, the identification of appropriate commonsense for response generation is tackled jointly with response generation itself, where a generative model's attention mechanisms implicitly learn which commonsense knowledge is most relevant to generating a response in each dialogue context \cite{zhang_conceptflow:20, sabour_cem:22, liu_commonsense:23}. 
However, recent work in a variety of text generation tasks suggests that breaking down a model into a sequence of explicit reasoning steps improves the correctness and overall quality of model outputs \cite{wei2022chain, chae_dialogue_2023, lee_mpc:23}.

Inspired by this, we hypothesize that dialogue modeling that modularizes commonsense reasoning into explicit steps will lead to more compelling dialogue responses. To explore this, we leverage Large Language Models (LLMs) to perform explicit commonsense reasoning for commonsense-augmented response generation since LLM-based approaches to dialogue have achieved remarkable success \cite{lee_gptempathy:22, kim_soda:22, chen_places:23}.
Our approach first generates multiple commonsense inferences for a dialogue context, covering several different social commonsense types. Since commonsense inferences can be produced that are true for a dialogue context but not useful for generating an appropriate response, a subset of the generated inferences are then explicitly selected based on their relevance to response generation. Finally, a response generator model produces a response for the dialogue context that integrates the selected commonsense. 

We find strong evidence that explicit reasoning over commonsense inferences leads to better dialogue interactions, improving the naturalness, engagingness, specificity, and overall quality of the responses against several alternative strategies for dialogue response generation.
Further analysis provides insights on which types of commonsense are most useful for response generation, revealing that predictions of personal characteristics and likely future events lead to the most favorable outcomes. Moreover, we assess human-provided explanations of response quality to identify the key response attributes that shape human perception of response quality and the particular response traits enhanced through commonsense integration. 
Taken together, our results demonstrate that isolating commonsense reasoning into explicit reasoning steps improves several aspects of response quality, achieving a new state-of-the-art in commonsense-augmented dialogue modeling. We make all code, models, and data publicly available to facilitate future work.\footnote{\url{https://github.com/emorynlp/DialCS-Reasoning}}

\begin{figure*}[t]
    \centering
    \includegraphics[width=\textwidth]{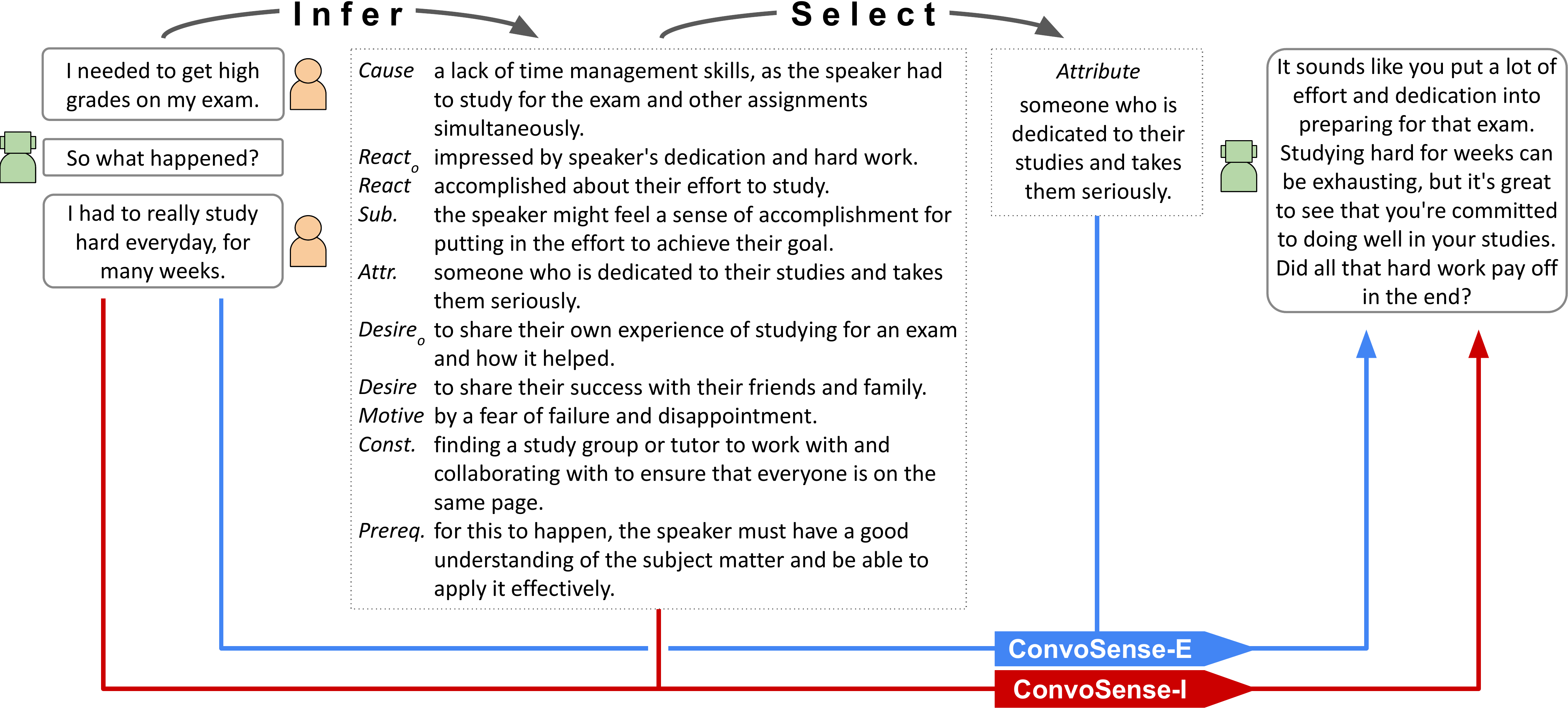}
    \caption{Overview of the explicit and implicit reasoning approaches (ConvoSense-\texttt{E/I}).}
    \label{fig:approach}
    \vspace{-2ex}
\end{figure*}

\section{Related Work}

Typical approaches to commonsense-augmented response generation include two steps, where a commonsense model first produces a set of commonsense inferences for a given dialogue context and then provides the inferences in conjunction with the dialogue context to a response generator model. Commonsense inference has been modeled as a retrieval process from a static commonsense knowledge base \cite{zhou_ccm:18, zhang_conceptflow:20, wu_conkadi:20, zhong_care:21, huang_stressor:22, wu_section:22, li_bridge:22, wu_channel:22, varshney_commonsense:22, li_ccf:23} such as ConceptNet \cite{speer_conceptnet:17} or using a commonsense generator that can produce novel commonsense inferences for a given context \cite{tu_misc:22, sabour_cem:22, fu_reasoning:23, liu_commonsense:23} such as COMET-ATOMIC \cite{bosselut_comet:19}. 
A limitation of these approaches is that the commonsense models focus on producing commonsense candidates that are true, regardless of their downstream appropriateness for response generation. Consequently, the response generator model must perform implicit commonsense reasoning to select which commonsense candidates should be integrated into the response. By contrast, our work hypothesizes that explicitly modeling which commonsense is appropriate to integrate into a response, independently of response generation, is a better strategy for augmenting a model with commonsense.

An alternative direction for commonsense-augmented response generation focuses on training a specialized commonsense generator with the goal of outputting only the commonsense which is relevant for generating responses in a given dialogue context. Some works accomplish this by training a single model to sequentially produce relevant commonsense for a dialogue context followed by a response that incorporates it \cite{liu_infused_2022, zhou_reflect_2022, zhou_tbs:22} whereas others train a specialized commonsense generative model to generate chain-of-thought sequences of commonsense inferences which are then used as input for response generation by LLMs \cite{chae_dialogue_2023}. 
A possible limitation of these specialized commonsense generators is that they jointly tackle two important reasoning steps together: determining true commonsense for a given dialogue situation and deciding on the relevance of specific commonsense for response generation. Our approach instead models these two steps explicitly using separate components. 

In summary, our work is the first to distinguish three explicit reasoning steps for commonsense-augmented response generation. Our method fully separates the tasks of commonsense generation, commonsense selection, and response generation, representing the most explicit approach for modeling dialogue commonsense to date.

\section{Response Generation via Explicit vs. Implicit Reasoning}

To explore the impact of explicit reasoning over commonsense inferences against the typically utilized implicit reasoning approach, we develop two prompt-based LLM strategies for response generation that treat generated inferences as speculative thoughts to guide follow-up response generation. These two approaches utilize the same inference generation procedure and differ on the strategy for integrating commonsense inferences into a follow-up response (Figure \ref{fig:approach}). The implicit reasoning variant provides all commonsense inferences as input to an LLM which is prompted to consider the inferences when generating the best follow-up response, similar to previous works. The explicit reasoning variant involves a three-step generate-select-and-respond procedure, using LLM prompting to explicitly identify the best commonsense inferences and subsequently synthesize them into a follow-up response. We first discuss the shared Inference Generation module (\S\ref{sec:inference-generation}), before detailing the implicit reasoning (\S\ref{sec:implicit-approach}) and the explicit reasoning approaches (\S\ref{sec:explicit-approach}).




\subsection{Inference Generation}
\label{sec:inference-generation}

The first step of both approaches is to identify multiple social commonsense inferences that are likely true for a given dialogue context. 


\paragraph{Inference Source}

To dynamically generate inferences relevant to a dialogue context, we use a generative model of social commonsense tailored for dialogue. Our aim is to produce individual inferences, which are reasonable to the dialogue context, are predictive in nature to the dialogue situation, and cover a wide variety of commonsense types. We adopt the ConvoSenseGenerator from \citet{finch_convosense_2024}, a T5-based model trained on the ConvoSense dataset to output inferences for a provided dialogue context and commonsense type. ConvoSenseGenerator excels in producing commonsense inferences across 10 different social commonsense types, surpassing existing works of ComFact \cite{gao:22}, Reflect \cite{zhou_reflect_2022}, and CICERO \cite{ghosal:22, shen:22} in type coverage as well as the reasonableness and predictiveness of the generated inferences. Furthermore, ConvoSenseGenerator allows precise control over the type of inferences generated, unlike the model proposed by \citet{chae_dialogue_2023} which outputs a set of three inferences with no explicit control over the outputted types.

\paragraph{Inference Distribution}

10 commonsense inferences are outputted for a provided dialogue context, each corresponding to one of the 10 commonsense types covered by ConvoSenseGenerator (Table \ref{tab:inference-prefixes}).  Initially, we explored outputting the top-ranked inference from beam search for each type but found significant semantic overlap in the inferences outputted across types. Since semantically unique inferences are critical for studying the impact of reasoning over these inferences, we implement a diverse beam search approach \cite{vijayakumar:16} to output five inferences per type and then select one inference per type such that between-type cosine similarity of SBERT embeddings \cite{reimers2019sentence} is minimized, following \citet{finch_convosense_2024} who show diverse beam search increases inference uniqueness.

\begin{table}[htbp]
    \centering
    \centering\resizebox{\columnwidth}{!}{
    \begin{tabular}{l|p{0.90\columnwidth}}
        \toprule
        \textbf{Type} & \textbf{Prefix} \\
        \midrule
        Cause & I think it is possible the previous dialogue turn was caused by \\
        React$_o$ & The Listener (You) feels \\
        React & I think the Speaker (Other) feels \\
        Subsequent & Next, I predict \\
        Attribute & I think the Speaker (Other) is \\
        Desire$_o$ & The Listener (You) wants \\
        Desire & I think the Speaker (Other) wants \\
        Motivation & I think the Speaker (Other) is motivated \\
        Constituent & I think it is possible the previous dialogue turn depends on \\
        Prerequisite & I think it is possible the previous dialogue turn requires \\
        
        \bottomrule
    \end{tabular}
    }
    \caption{Prefixes used for the commonsense inferences.}
    \label{tab:inference-prefixes}
    \vspace{-2ex}
\end{table}


\paragraph{Inference Representation}

Inferences generated by ConvoSenseGenerator are augmented with natural language prefixes, transforming them into complete sentences. These prefixes serve to indicate the level of speculation inherent in the predictions. Inferences pertaining directly to the conversational role played by the system are treated as factual, while those concerning the other interlocutor in the conversation or the dialogue situation itself are considered speculative. Table \ref{tab:inference-prefixes} provides the ten inference types and their corresponding prefixes.


\begin{table}[htb]
    \centering
    \small
    \begin{tabular}{@{}p{\columnwidth}@{}}
        \toprule
        \begin{tabular}[t]{@{}p{\columnwidth}@{}}
            You are the Listener in a conversation shown in \q{Dialogue History}.\\
            \\
            Your goal is write a casual yet engaging and appropriate next response for the Listener (You) in the provided dialogue. You will consider a list of possible \q{Talking Points} to include as you think about the best response to give, being careful to ignore any talking points that are irrelevant or unlikely predictions for the shown conversation. \\
            \\
            Based on the talking points, write the best response you can think of in the following format: \\
            \\
            Listener's Response: \\
            \_\_\_ \\
            \\
            Review the following examples to understand how to write a response given a \q{Dialogue History} and set of possible \q{Talking Points}.\\
            \\
            \{examples\}\\
            \\
            Now, construct the best response from the Listener for the following dialogue, based on the possible talking points:\\
            \\
            \# Dialogue History\\
            \{context\}\\
            \\
            \# Talking Points\\
            \{inferences\}\\
            \\
            Listener's Response:
        \end{tabular}  \\
        \bottomrule
    \end{tabular}
    \caption{The prompt used for ConvoSense-\texttt{I}. \texttt{\{context\}}, \texttt{\{inferences\}}, and \texttt{\{examples\}} are filled by dialogue context, commonsense, and few-shots.}
    \label{tab:crest-implicit-prompt}
    \vspace{-2ex}
\end{table}

\subsection{Implicit Reasoning}
\label{sec:implicit-approach}

Given the output from inference generation, the implicit reasoning approach immediately performs response generation by taking all generated inferences as input.
Table~\ref{tab:crest-implicit-prompt} provides the prompt for this approach, referred to as ConvoSense-\texttt{I}.

\paragraph{Response Generation}

The Response Generation module takes as input the set of generated inferences and the dialogue context, and outputs the next response.  We use GPT-3.5 for response generation, which is instructed to carefully consider all of the commonsense inferences and then write the best response based on this consideration. This produces a dialogue response that is grounded on implicitly selected commonsense inferences.

\begin{table}[htb!]
    \centering
    \small
    \resizebox{0.95\columnwidth}{!}{%
    \begin{tabular}{@{}p{\columnwidth}@{}}
        \toprule
        \begin{tabular}[t]{@{}p{0.98\columnwidth}@{}}
            You find yourself in the role of a conversational architect, who is responsible for setting up the next exchange in the ongoing dialogue presented in \q{Dialogue History.} Specifically, your task is to review the series of talking points provided in \q{Talking Points} and select the best 1 idea that will craft an engaging and cohesive response for the Listener to say. Write your selected talking point into a list titled \q{Selection}. \\ 
            \\
            Review the following examples of good selections for different pairs of \q{Dialogue History} and \q{Talking Points}.\\
            \\
            \{examples\} \\
            \\
            Now, select the best talking point for the following pair: \\
            \\
            \# Dialogue History \\
            \{context\} \\
            \\
            \# Talking Points \\
            \{inferences\} \\
            \\
            Selection:
        \end{tabular}  \\
        \midrule
        \begin{tabular}[t]{@{}p{0.98\columnwidth}@{}}
            You are the Listener in a conversation shown in \q{Dialogue History}. \\
            \\
            Your goal is write a casual yet engaging and appropriate next response for the Listener (You) in the provided dialogue. First, sufficiently answer all questions posed by Speaker (Other) in their preceding turn. Then, continue your response by including the talking points shown in \q{Talking Points} since you want to cover them in your next response too. \\
            \\
            Write the response in the following format: \\
            \\
            Listener's Response:\\
            \_\_\_\\
            \\
            Review the following examples to understand how to write a response given a \q{Dialogue History} and set of \q{Talking Points}.\\
            \\
            \{examples\}\\
            \\
            Now, complete the tasks for the following situation:\\
            \\
            \# Dialogue History \\
            \{context\} \\
            \\
            \# Talking Points \\
            \{inferences\} \\
            \\
            Listener's Response: \\
        \end{tabular} \\
    \bottomrule
    \end{tabular} }
    \caption{The Inference Selection (top) and Response Generation (bottom) prompts of ConvoSense-\texttt{E}. \texttt{\{context\}}, \texttt{\{inferences\}}, and \texttt{\{examples\}} are filled by dialogue context, commonsense, and few-shots.}
    \label{tab:crest-explicit-prompt}
    \vspace{-2ex}
\end{table}

\subsection{Explicit Reasoning}
\label{sec:explicit-approach}

Given the output from inference generation, the explicit reasoning approach selects relevant inferences before composing the follow-up response. Table \ref{tab:crest-explicit-prompt} provides the prompts for this approach, referred to as ConvoSense-\texttt{E}.

\paragraph{Inference Selection} 

The goal of inference selection is to identify which inferences are most useful towards generating an interesting and appropriate response to the dialogue context. GPT-3.5 is tasked with selecting $k$ inferences from the full set of inferences by being prompted to carefully consider each inference and strategically determine which inferences are the most useful, relevant, and interesting for the next response in the dialogue context. The selected inferences are outputted as a list.

\noindent
The determination of the number $k$ of inferences to select is treated as a hyperparameter to be optimized. In pilot studies, we observed that $k=1$ performed best, since increasing $k$ often resulted in longwinded, unfocused responses that integrated too many disparate commonsense inferences.


\paragraph{Response Generation}

After inference selection, response generation takes as input the list of selected inferences and the dialogue context, and outputs the next response. GPT-3.5 is instructed to synthesize the semantic content provided in the selected inferences into an engaging and appropriate response, producing a dialogue response grounded on explicitly selected commonsense inferences.

\subsection{Prompt Formatting}

For all GPT prompts, the dialogue context is provided as a sequence of turns, prefixed with speaker labels. 10 few-shot examples are provided to both ConvoSense-\texttt{E} and ConvoSense-\texttt{I}. For ConvoSense-\texttt{E}, we construct 10 inference selection examples, one for each inference type included in this study, and 100 response generation examples, 10 for each inference type. During inference, the 10 inference selection examples are used in the inference selection prompt. After inference selection, then the 10 response generation examples corresponding to the selected inference type are used in the response generation prompt. For ConvoSense-\texttt{I}, we choose 10 response generation examples from those crafted for ConvoSense-\texttt{E}, ensuring one example for each commonsense type. During inference, these 10 response generation examples are used in the response generation prompt.

\section{Experiments}

To study the impact of explicit reasoning over commonsense inferences on response generation, we compare ConvoSense-\texttt{E} against three alternative approaches, two which utilize implicit reasoning over commonsense (ConvoSense-\texttt{I} and Doctor) and one without access to external commonsense (GPT).

\paragraph{ConvoSense-\texttt{I}} represents the implicit reasoning approach that is a direct comparison against ConvoSense-\texttt{E} (Section \ref{sec:implicit-approach}).

\paragraph{Doctor} is the state-of-the-art for commonsense-augmented dialogue. It uses an implicit reasoning approach in which a trained commonsense model generates a subset of commonsense types for a given dialogue context that are then provided to GPT-3.5 for response generation. We use the released model and prompt from \citet{chae_dialogue_2023}.

\paragraph{GPT} is a model representing the baseline capability of GPT-3.5 for dialogue response generation. No commonsense or commonsense-derived response examples are provided in order to elicit the natural tendencies of GPT. The prompt is shown in Table \ref{tab:native-chatgpt}.

\begin{table}[htb]
    \centering
    \small
    \begin{tabular}{@{}p{\columnwidth}@{}}
        \toprule
        \begin{tabular}[t]{@{}p{\columnwidth}@{}}
            \# Dialogue History \\
            \{context\} \\
            \\
            You are the Listener in a conversation shown in \q{Dialogue History}. \\
            \\
            Your goal is write a casual yet engaging and appropriate next response for the Listener (You) in the provided dialogue. \\
            \\
            Write the response in the following format:\\
            \\
            Listener's Response:\\
            \_\_\_\\
            \\
            Listener's Response: \\
        \end{tabular}  \\
        \bottomrule
    \end{tabular}
    \caption{The prompt used for native response generation of GPT. \texttt{\{context\}} is filled by the dialogue context of a provided example.}
    \label{tab:native-chatgpt}
    \vspace{-2ex}
\end{table}


\vspace{2ex}
\noindent We employ the same GPT-3.5 version in each approach, choosing the latest version at the time (\texttt{gpt-3.5-turbo-0125}) and temperature of $0.7$.

\subsection{Test Data}

To conduct a realistic evaluation of the approaches under study, the test data consists of dialogues sampled from an \q{out-of-distribution} dataset for all models: Reflect \cite{zhou_reflect_2022}. The Reflect dataset is composed of human-written dialogues that are based on descriptions of everyday situations.
100 dialogues are sampled for use in the evaluations, with an average of 3.1 turns and 10.8 words per utterance.

\subsection{Evaluation}
\label{sec:subjective_preferences_eval}

\begin{figure}[htb]
    \centering
    \includegraphics[width=\columnwidth]{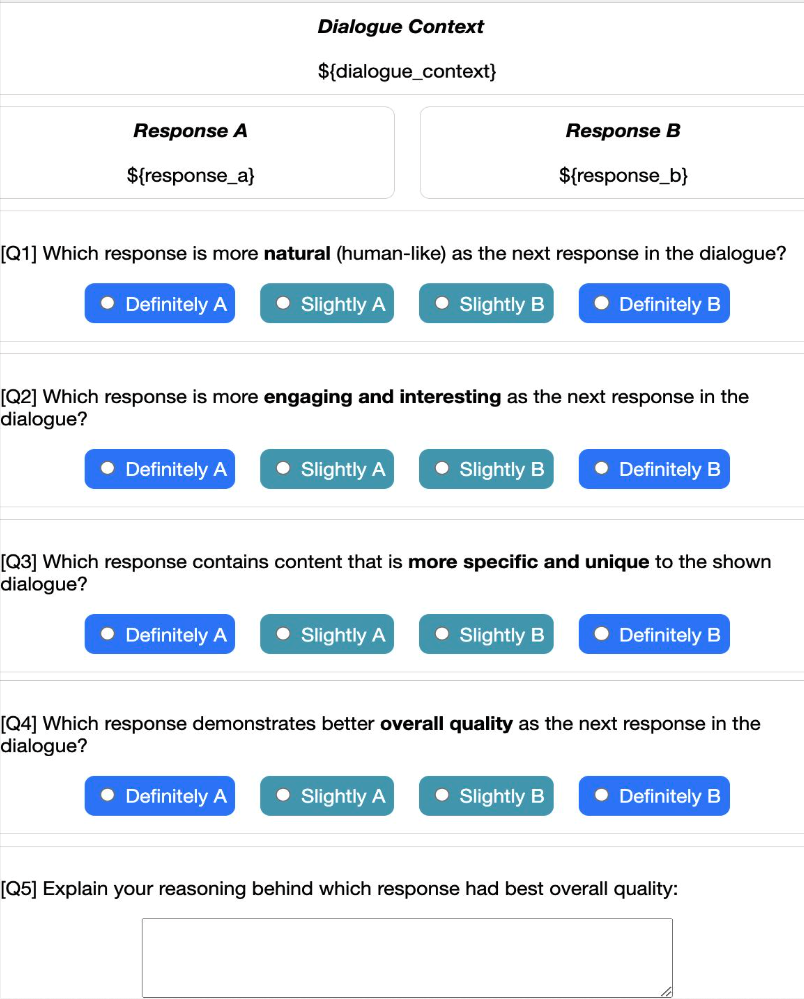}
    \caption{Evaluation interface.}
    \label{fig:mturk-interface}
    \vspace{-2ex}
\end{figure}

To capture differences in end-user impressions of the different dialogue approaches, we perform pairwise preference selections between pairs of dialogue approaches using human judges. Responses from two different dialogue approaches for the same dialogue context are shown to human judges who are instructed to identify which response better satisfies the indicated characteristic. Following the improvements in dialogue characteristics from earlier commonsense-augmented dialogue works \cite{zhou_reflect_2022, chae_dialogue_2023}, we evaluate the response characteristics shown in Figure \ref{fig:mturk-interface}: Naturalness (Q1), Engagingness (Q2), Specificity (Q3), and Overall Quality (Q4).
Human judges are also instructed to provide a freeform text explanation of their reasoning behind which response had the better quality. 

Three Amazon Mechanical Turkers perform each evaluation task and are paid \$0.15 USD per task. Our pilot studies on MTurk reveal that standard worker filtering criteria (Location: USA,CAN; HIT approval $\ge$ 98\%; approved HITs: $\ge$ 10000) fail to ensure good-faith workers for our evaluation tasks, often yielding invalid or nonsensical explanations. To address this, we introduce a \q{screening} task identical to the evaluation task, used to identify reliable workers based on their written explanations. This process identifies over thirty MTurkers who provide detailed explanations for their responses. These selected workers then perform the evaluation tasks for our dialogue approaches and achieve moderate inter-annotator agreement, with Krippendorff's alpha scores of 0.442, 0.560, 0.595, and 0.561 for naturalness, engagingness, specificity, and quality, respectively.

\begin{table*}[htb]
    \centering
    \footnotesize
    \resizebox{\textwidth}{!}{%
    \begin{tabular}[t]{@{}p{0.07\textwidth}|p{0.93\textwidth}}
        \multirow{3}{*}{DC} & 
        \begin{tabular}[t]{rp{\textwidth}}
            Speaker (Other): & I had to kick Jan out of my house last night. \\
            Listener (You): & What got you so mad that you kicked her out of the house? \\
            Speaker (Other): & She kept bugging the dog and bothering him. \\
        \end{tabular} 
        \\
        \midrule
        \midrule
        GPT & That's not cool, dogs need their space too. Did she understand why you were upset with her behavior?
        \\
        \midrule
        Doctor &
        \begin{tabular}[t]{@{}p{0.89\textwidth}@{}}
            That's not a good reason to kick someone out.
        \end{tabular} 
        \\
        \midrule
        \multirow{2}{*}{CS-\texttt{I}} & \begin{tabular}[t]{@{}p{0.89\textwidth}@{}}
            That must have been a tough situation. I hope the dog is okay. Do you think it's time for Jan to find a new place to stay?
        \end{tabular}
        \\
        \midrule
        \multirow{2}{*}{CS-\texttt{E}} & \begin{tabular}[t]{@{}p{0.89\textwidth}@{}}
            I can understand why you'd be upset if Jan was bothering your dog. Your home should be a peaceful and safe space for everyone, including your furry friend. How is the dog doing now that Jan's not around?
        \end{tabular}
    \end{tabular}%
    }
    \caption{Example response outputs for a dialogue context (DC) from each model. CS-\texttt{E/I} denote ConvoSense-\texttt{E/I}.}
    \label{tab:output-examples-brief}
    \vspace{-2ex}
\end{table*}

\subsection{Results}

\begin{table}[htb]
    \centering
    \resizebox{\columnwidth}{!}{%
    \begin{tabular}{r|rrrr}
        \toprule 
        & Natural & Engaging & Specific & Quality \\
        \toprule
        ConvoSense-\texttt{E} & \bf 82.7 & \bf 92.3 & \bf 91.3 & \bf 92.0 \\
        Doctor & 17.3 & 7.7 & 8.7 & 8.0 \\
        \midrule
        ConvoSense-\texttt{E} & \bf 75.7 & \bf 82.7 & \bf 86.3 & \bf 84.3 \\
        GPT & 24.3 & 17.3 & 13.7 & 15.7 \\
        \midrule
        ConvoSense-\texttt{E} & \bf \underline{55.3} & \bf 66.7 & \bf 63.7 & \bf 63.7 \\
        ConvoSense-\texttt{I} & 44.7 & 33.3 & 36.3 & 36.3 \\
        \midrule
        ConvoSense-\texttt{I} & \bf 84.7 & \bf 89.3 & \bf 86.3 & \bf 89.7 \\
        Doctor & 15.3 & 10.7 & 12.7 & 10.3 \\
        \midrule
        ConvoSense-\texttt{I} & \bf 68.7 & \bf 67.7 & \bf 73.0 & \bf 70.3 \\
        GPT & 31.3 & 32.3 & 27.0 & 29.7 \\
        \bottomrule 
    \end{tabular} }
    \caption{Pairwise evaluation results showing the preference percentages. Winning models in each comparison are statistically significant for all characteristics, except where \underline{underlined} (proportion test, $p < 0.01$).} 
    \label{tab:pairwise-comparisons} 
    \vspace{-2ex}
\end{table}

\noindent Table \ref{tab:output-examples-brief} presents examples of the response outputs from each model\footnote{Full examples including commonsense generations can be found in Appx. \ref{app:output-examples}.} and Table \ref{tab:pairwise-comparisons} shows the preference results for the models based on the human evaluation. These evaluation results overwhelmingly indicate that explicit reasoning over dialogue-relevant commonsense inferences results in more appealing responses to human judges. This approach consistently generates responses that are preferred by human judges for their engagingness, specificity, and overall quality, although the naturalness of the follow-up response is least affected by the choice of commonsense reasoning strategy. When directly comparing explicit to implicit reasoning over the same commonsense inferences, as elucidated by the pairwise selections between ConvoSense-\texttt{E} and ConvoSense-\texttt{I}, there is not a strong difference with regard to response naturalness. This suggests that, although explicit reasoning aids in the engagingness and specificity of the response to its dialogue context, it is not as useful towards improving how natural the response is for the dialogue. Overall, in this era of leveraging powerful LLMs to perform tasks with little-to-no direct training, work like that in this study is revealing that modularized generation strategies can outperform end-to-end modeling paradigms that have dominated in the past. 

Furthermore, ConvoSense-\texttt{I} is found to be the most competitive approach to ConvoSense-\texttt{E}, with the highest rate of preference wins relative to the alternative approaches. This is further corroborated by direct pairwise comparisons between ConvoSense-\texttt{I} and each of the alternative approaches, Doctor and GPT, although the preference is not as strong as that for ConvoSense-\texttt{E}. 
Surprisingly, we find that the Doctor responses are the least preferred, even failing to outperform GPT itself, and we explore this further in Section \ref{sec:prompt-variation}.
Regardless, the strong performance of both ConvoSense-\texttt{E} and ConvoSense-\texttt{I} indicate that augmenting responses with appropriate commonsense improves on the native response generation of GPT.

\begin{figure*}[htb!]
    \centering
    \includegraphics[width=\textwidth]{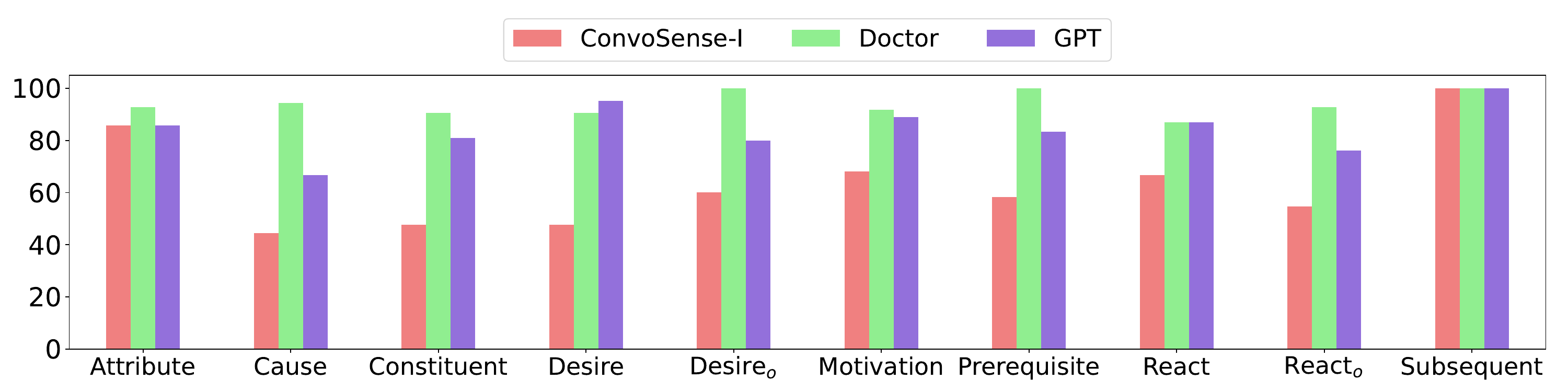}
    \caption{Percentage that ConvoSense-\texttt{E} wins against other models on Quality, split by the type of commonsense selected for response integration by ConvoSense-\texttt{E}.}
    \label{fig:results-by-cstype}
    \vspace{-2ex}
\end{figure*}

\section{Discussion}

\subsection{Impact of Commonsense Source} 

As discussed in Section \ref{sec:inference-generation}, there are several sources for commonsense for dialogue. Through the experiments in this study, we are able to compare the downstream utility of two of these sources: that used in Doctor, which has been shown to lead to better response outcomes than the resources that came before it, and ConvoSense, which has not been applied to response generation before this study.  Specifically, the pairwise selection between ConvoSense-\texttt{I} and Doctor enables a comparison of the underlying commonsense resources since their main difference is the source of commonsense inferences used in their approaches. The strong preference for ConvoSense-\texttt{I} against Doctor demonstrates the superiority of the ConvoSense inferences for leading to compelling responses in dialogue. This provides evidence that the commonsense resource used in this work advances dialogue modeling beyond the resource used in the previous state-of-the-art model, Doctor.

\subsection{Impact of Commonsense Type}

As indicated in \citet{zhou_reflect_2022}, the choice of grounding commonsense type has an impact on the resulting quality of the response for a given dialogue context. To better understand the effect of each commonsense type on response generation in the context of explicit commonsense reasoning, we decompose the pairwise selection results into isolated results for each type of selected commonsense. Figure \ref{fig:results-by-cstype} shows the pairwise selection results between ConvoSense-\texttt{E} and the other approaches, split into groupings of test instances based on the type of commonsense selected by ConvoSense-\texttt{E}. 

From these results, it can be seen that responses that integrate \texttt{Attribute} and \texttt{Subsequent} commonsense inferences consistently perform quite well against the alternative approaches. On the other hand, responses that integrate \texttt{Cause}, \texttt{Constituent}, \texttt{Desire}, and \texttt{React$_o$} inferences seem to perform the worst, especially against ConvoSense-\texttt{I}. 
These decomposed results reveal potential weaknesses of GPT-3.5 to reasoning and generation on certain commonsense types for response generation (e.g.~\texttt{Cause}, \texttt{Constituent}, \texttt{Desire}, and \texttt{React$_o$}). 
The approach used in this work relies on the capability of GPT-3.5 to reason about and synthesize commonsense inferences for response generation given a handful of appropriate few-shot examples.
Further work on explicit reasoning processes for commonsense-augmented dialogue should explore improvements to the reasoning process and integration strategies to ensure each commonsense type is utilized optimally.

    
    

\begin{figure*}[htb]
    \centering
    \includegraphics[width=\textwidth]{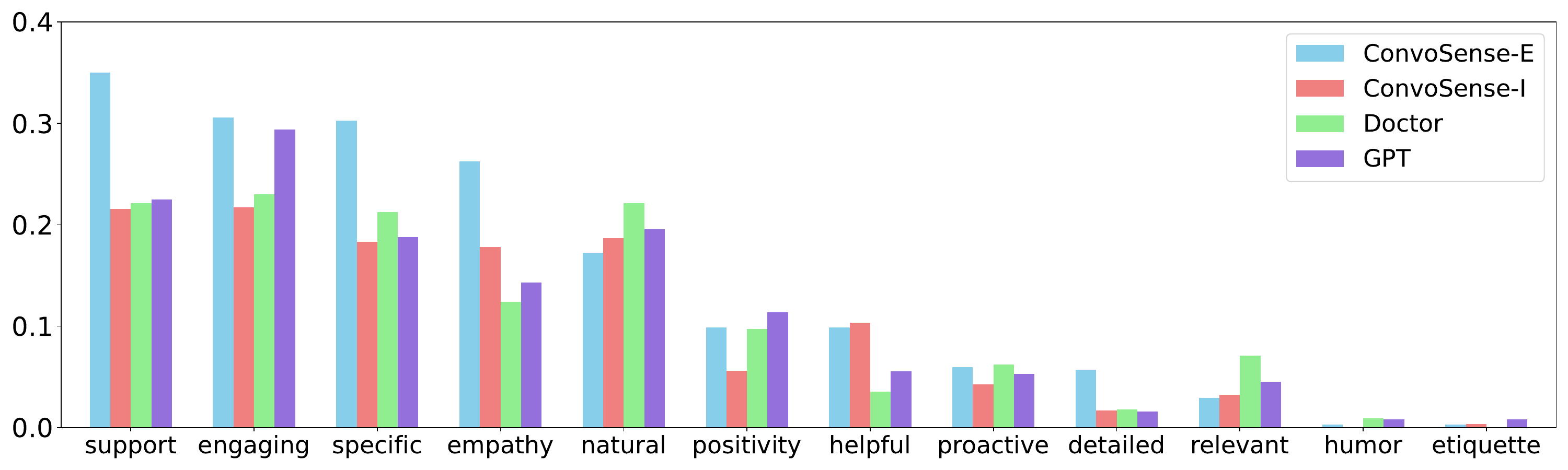}
    \caption{Proportion of explanations that include each response characteristic for each model.}
    \label{fig:modelwise-aspects}
    \vspace{-2ex}
\end{figure*}

\subsection{Influential Aspects from Human Feedback}
\label{sec:influential-aspects}

Based on the preference results in Table \ref{tab:pairwise-comparisons}, the overall quality preference is more aligned with the judgements on which response is more engaging and specific to the context, and less aligned with the judgements for naturalness. To explore the influential aspects on perceived response quality in greater detail, we examine the textual explanations of overall quality preference provided by the human judges for each of their judgements. 

We utilize an automated aspect identification procedure to summarize the human-written explanations into sets of influential characteristics of the preferred response. This aspect identification procedure uses GPT-3.5 to output short phrases (one or two words) that represent each of the indicated characteristics mentioned in the explanation (Table~\ref{tab:automated-aspect-prompt}; Appx. \ref{app:influential-aspects}). On an example set of 50 explanations with characteristics identified by a human annotator, this procedure obtains 0.82 precision and 0.83 recall of the characteristics. 

We run this aspect identification procedure on the explanations collected from the evaluations. We then manually review the outputted individual characteristics and find that there is a high degree of distinct yet synonymous characteristics being outputted. To aggregate the characteristics by synonymous meaning, we construct a mapping between the outputted characteristics to 12 categories. Table \ref{tab:output-examples} (Appx. \ref{app:output-examples}) shows examples of the predicted aspects and their corresponding mapped categories. Figure \ref{fig:modelwise-aspects} shows the distribution of categories in the explanations for each winning model.


Across all models, there are five characteristics that are most often cited as a determining factor for the identification of a better response: the supportive nature of the response, the level of engaging and interesting information, the specificity of the response contents, the degree of empathy, and the naturalness. This suggests that these characteristics are the most influential aspects contributing to response favorability among human judges.


Specifically for ConvoSense-\texttt{E}, it has a higher rate of explanations that highlight the response specificity, supportiveness, and detailedness as influential features for overall response preference, compared to the other models. This showcases that the ConvoSense-\texttt{E} approach is more capable than the other models at producing specific, supportive, and detailed responses to a provided dialogue context. In addition, it can be seen that both approaches that utilize ConvoSense inferences receive more preference wins due to the empathetic and helpful nature of their responses, suggesting that the commonsense used in these approaches is useful for improving the empathy displayed in the responses and encouraging the responses to be solution/advice-oriented to the events being discussed in the dialogue, more so than that provided by responses from Doctor or native GPT.

\subsection{Baseline Prompt Variation} 
\label{sec:prompt-variation}


One unexpected outcome of this work is the poor performance of the Doctor approach. As noted in Section \ref{sec:subjective_preferences_eval}, Doctor is the least competitive approach against ConvoSense-\texttt{E} across all of the characteristics. This is contrary to the results presented in \citet{chae_dialogue_2023}, which indicate that Doctor outperforms GPT in terms of naturalness and specificity. To explicitly compare the GPT baseline used in our work against Doctor, we conduct the pairwise preference selection evaluation between them. From the results in Table \ref{tab:chatgpt-doctor-pairwise-comparisons}, it is shown that Doctor is indeed outperformed by our GPT baseline. 

\begin{table}[htb]
    \centering
    \resizebox{\columnwidth}{!}{%
    \begin{tabular}{rc|c|c|c}
        \toprule 
        & \multicolumn{1}{c}{Natural} & \multicolumn{1}{c}{Engaging} & \multicolumn{1}{c}{Specific} & \multicolumn{1}{c}{Quality} \\
        \toprule
        GPT & \bf 72.0 & \bf 82.3 & \bf 77.7 & \bf 80.7 \\
        Doctor & 28.0 & 17.7 & 22.3 & 19.3 \\
        \bottomrule 
    \end{tabular} }
    \caption{Pairwise evaluation results showing the win percentages. All wins are statistically significant (proportion test, $p < 0.01$).} 
    \label{tab:chatgpt-doctor-pairwise-comparisons} 
    \vspace{-1ex}
\end{table}

\noindent We hypothesize that we have utilized a stronger GPT baseline in this work than that used in \citet{chae_dialogue_2023}.  
To verify this hypothesis, we conduct the pairwise preference selection evaluation between responses from the GPT approach from this work and the prompt released by \citet{chae_dialogue_2023}. 
From the results in Table \ref{tab:chatgpt-comparisons}, we are able to confirm that our GPT prompt does indeed lead to stronger responses than that used previously, thus helping to explain the evaluation discrepancies observed in this work. This difference in outcomes between two different GPT prompts highlights the need for careful construction of prompts for using LLM capabilities as baselines, in order to ensure appropriate representation of the power of the LLM and to avoid an overestimate of the impact of new dialogue approaches. It is also possible that GPT performance has improved in general since the publication of \citet{chae_dialogue_2023}, further emphasizing the need to continue to include such baseline models in follow-up experiments to track performance progression.

\begin{table}[htb]
    \centering
    \resizebox{\columnwidth}{!}{%
    \begin{tabular}{lc|c|c|c}
        \toprule 
        & \multicolumn{1}{c}{Natural} & \multicolumn{1}{c}{Engaging} & \multicolumn{1}{c}{Specific} & \multicolumn{1}{c}{Quality} \\
        \toprule
        GPT & \bf \underline{55.0} & \bf 58.0 & \bf 57.7 & \bf 61.3 \\
        GPT$_{chae}$ & 45.0 & 42.0 & 42.3 & 38.7 \\
        \bottomrule 
    \end{tabular} }
    \caption{Pairwise evaluation results showing the win percentages. All wins are statistically significant, except where \underline{underlined} (proportion test, $p < 0.01$).} 
    \label{tab:chatgpt-comparisons} 
    \vspace{-3ex}
\end{table}

\section{Conclusion}

The findings of this paper underscore the benefits of an explicit approach for incorporating commonsense into dialogue responses, in which the separate generation, selection, and integration of commonsense into dialogue responses enables improvements in response quality. Our findings not only showcase the efficacy of this explicit reasoning model but also shed light on the types of commonsense most beneficial for response generation and reveal the fine-grained response characteristics that are improved through this explicit reasoning process. By elucidating these advancements and insights, we contribute to the ongoing evolution of dialogue systems, making interactions more engaging, contextually aware, and satisfying for users. We anticipate future research will extend the scope of explicit reasoning in dialogue response generation to encompass a broader range of information sources and alternative reasoning strategies, thus supporting the advancement of tailoring dialogue systems to diverse domains and user populations.

\clearpage
\section{Limitations}

By showcasing the advantage of explicit reasoning steps for commonsense integration into dialogue response generation, the findings of our work provide valuable insights into a fruitful direction for future dialogue works to follow to optimize dialogue performance. To further understand the impact of this explicit reasoning and identify any outstanding challenges, there are a few limitations to be noted in our work that can inform follow-up investigations. 



\paragraph{Generality Beyond GPT-based Systems} Our experiments primarily focused on GPT-3.5-based dialogue systems. Although we attempted to extend our methods to Llama2, we observed poor performance. This suggests that additional work is necessary to implement explicit reasoning to other, less powerful models. Future work should explore the implementation of explicit reasoning for a broader range of models and investigate fine-tuning approaches to enhance performance across different LLMs.

\paragraph{Strategy of Explicit Reasoning} The explicit reasoning step undertaken in our study involves selecting a single commonsense inference from a large pool of candidates, which will be integrated into the follow-up response. However, there are many alternative reasoning strategies that could be explored. For instance, strategies could be developed to prioritize inferences that address the user's emotional needs, generate intelligent follow-up questions, or achieve other specific dialogue goals. Investigating these alternative strategies could reveal further enhancements in response quality and user engagement depending on the dialogue application.

\paragraph{Commonsense Information Source} Our research investigates explicit reasoning specifically in the context of social commonsense inferences. Future work should explore expansion to additional types of commonsense, such as temporal or property-based, to further the investigation of commonsense-augmented dialogue models and the utility of explicit reasoning. 

\paragraph{Static Evaluation} We follow the evaluation paradigm of previous commonsense-augmented dialogue works \cite{zhou_reflect_2022, chae_dialogue_2023} in which a response is generated for a static dialogue context. Although this provides an understanding of the response generation capabilities of dialogue models, real-world deployment of such systems that involves multi-turn back-and-forth interactions can reveal aspects of dialogue models that are not demonstrated through static evaluations. Future work should explore the deployment of dialogue systems with explicit reasoning over commonsense to further understand their performance with human users.


\section{Ethical Considerations}

\paragraph{Bias and Stereotyping}
One important consideration regarding integrating commonsense reasoning into dialogue response generation is the potential for perpetuating stereotypes due to the generalized nature of commonsense knowledge. This could result in dialogue systems producing responses that reflect these stereotypes or exhibit unfair biases. While we would expect there to be a negative impact on human reception of these responses if they are significantly biased, which is not observed in this study, it is possible that human evaluators share similar stereotypes or biases and therefore do not find these responses uncomfortable. This highlights the importance of future research to thoroughly investigate the risks of bias in commonsense reasoning for response generation, ensuring the development of equitable AI systems.


\paragraph{Risks of Explicit Dialogue Model Control}
Having a dialogue system design that relies on explicit reasoning steps enables the opportunity for antisocial reasoning or response strategies to be directly inserted into a model, which can lead to a higher rate of such behaviors being expressed as compared to indirect learning from training data. At the same time, however, it also affords opportunities to promote strategies that aim to reduce such antisocial response behaviors. This controllable approach contrasts with end-to-end dialogue systems, providing a more precise method for mitigating harmful outputs, which we leave to future work to explore the success of such strategies.

\paragraph{Compensation of Human Evaluators}
We ensure fair compensation for the human evaluators involved in our study, with an estimated hourly pay rate of \$12 USD, which exceeds minimum wage.

\bibliography{custom}

\clearpage
\appendix


\section{Output Examples}
\label{app:output-examples}

Full examples of the response outputs, commonsense, evaluation explanations, predicted aspects, and mapped categories are shown in Table \ref{tab:output-examples} for the models under study.

\section{Automated Aspect Identification}
\label{app:influential-aspects}

Table \ref{tab:automated-aspect-prompt} shows the prompt for automated aspect identification in Section \ref{sec:influential-aspects}. The collected human explanations from the evaluation are segmented into batches of 10 and are passed to GPT-3.5 using this prompt.

\begin{table}[htb]
    \centering
    \small
    \begin{tabular}{@{}p{\columnwidth}@{}}
        \toprule
        \begin{tabular}[t]{@{}p{\columnwidth}@{}}
            I have received feedback from human judges explaining their preference for a certain dialogue response from the options displayed to them. For each of the following explanations, please list the positive aspects identified. Aspects should be one word only, so please summarize the positive traits identified into one word if needed. Examples of aspects that could be mentioned are empathy, engagement, curiosity, acknowledgement, support, naturalness, and more. \\
            \\
            Output a list of aspects for each explanation below. \\
            \\
            \{explanations\} \\
        \end{tabular}  \\
        \bottomrule
    \end{tabular}
    \caption{The prompt used for automated aspect identification of human preference explanations. \texttt{\{explanations\}} is filled with human explanations as a numbered list.}
    \label{tab:automated-aspect-prompt}
\end{table}

\begin{table*}[htb]
    \centering
    \footnotesize
    \resizebox{\textwidth}{!}{%
    \begin{tabular}[t]{@{}p{0.07\textwidth}|p{0.93\textwidth}}
        \multirow{3}{*}{DC} & 
        \begin{tabular}[t]{llrp{\textwidth}}
            Speaker (Other) & / & B: & I had to kick Jan out of my house last night. \\
            Listener (You) & / & A: & What got you so mad that you kicked her out of the house? \\
            Speaker (Other) & / & B: & She kept bugging the dog and bothering him. \\
        \end{tabular} 
        \\
        \midrule
        GPT & That's not cool, dogs need their space too. Did she understand why you were upset with her behavior?
        \\
        \midrule
        \multirow{9}{*}{Doctor} &
        \begin{tabular}[t]{@{}p{0.89\textwidth}@{}}
            Subquestion 1: What is the reason for Person B's anger towards their neighbor? (xAttr) \\
            Subanswer 1: Person B is angry because their neighbor kept bugging the dog and bothering him. \\
            Subquestion 2: How does Person A perceive Person B's behavior towards their neighbor? (oReact) \\
            Subanswer 2: Person A perceives Person B's behavior as being childish and not helpful towards their neighbor. \\
            Subquestion 3: What might Person A want to communicate to Person B, given their perception of Person B's behavior? (oWant) \\
            Subanswer 3: Person A might want to express their disapproval of Person B's behavior and suggest that it is not acceptable. \\
            \\
            That's not a good reason to kick someone out.
        \end{tabular} 
        \\
        \midrule
        \multirow{18}{*}{CS-\texttt{I}} & \begin{tabular}[t]{@{}p{0.89\textwidth}@{}}
            $\ast$ I think the Speaker (Other) feels guilty for having to deal with jan's behavior. \\
            $\ast$ The Listener (You) feels concerned about the well-being of the dog and wonders if there are any underlying issues that could be causing the behavior. \\
            $\ast$ Next, I predict the listener might ask the speaker if they have any other pets in the house. \\
            $\ast$ I think it is possible the previous dialogue turn depends on jan not respecting the boundaries of the house and not being respectful of the speaker's property. \\
            $\ast$ The Listener (You) wants to express sympathy for speaker's situation and offer to help him find a new living situation. \\
            $\ast$ I think it is possible the previous dialogue turn requires jan had access to the dog's living space. \\
            $\ast$ I think it is possible the previous dialogue turn was caused by the dog being a rescue dog and jan being a new owner. \\
            $\ast$ I think the Speaker (Other) wants to find a new place to live that is more peaceful and doesn't have any pets. \\
            $\ast$ I think the Speaker (Other) is someone who takes their pets seriously and doesn't tolerate any behavior that could harm them. \\
            $\ast$ I think the Speaker (Other) is motivated by a need for peace and quiet in their home. \\
            \\
            That must have been a tough situation. I hope the dog is okay. Do you think it's time for Jan to find a new place to stay?
        \end{tabular}
        \\
        \midrule
        \multirow{4}{*}{CS-\texttt{E}} & \begin{tabular}[t]{@{}p{0.89\textwidth}@{}}
            $\ast$ I think the Speaker (Other) is motivated by a need for peace and quiet in their home. \\
            \\
            I can understand why you'd be upset if Jan was bothering your dog. Your home should be a peaceful and safe space for everyone, including your furry friend. How is the dog doing now that Jan's not around?
        \end{tabular}
        \\
        \midrule 
        \midrule
        \multirow{12}{*}{EX} & \begin{tabular}[t]{lp{0.50\textwidth}l}
            vs. GPT & Response A is better overall choice.it shows empathy towards speaker 1's situation,acknowledges the importance of a peaceful environment for both humans and animals ,and expresses concern for the well-being os speaker 1's dog. & \begin{tabular}[t]{p{0.10\textwidth}l}
                \textit{empathy} & \textbf{empathy} \\
                \textit{acknowledge} & \textbf{specific} \\
                \textit{concern} & \textbf{support}
                \end{tabular}  
            \\
            \midrule
            vs. Doctor & Response B is better as it shows more concern, expresses understanding and empathy for their situation. & \begin{tabular}[t]{p{0.10\textwidth}l}
                \textit{concern} & \textbf{support} \\
                \textit{understanding} & \textbf{support} \\
                \textit{empathy} & \textbf{empathy}
                \end{tabular}  
            \\
            \midrule
            vs. ConvoSense-\texttt{I} & The given response is more relevance to the conversation and make more comprehensive & \begin{tabular}[t]{p{0.10\textwidth}l}
                \textit{relevance} & \textbf{relevant} \\
                \textit{comprehensive} & \textbf{detailed}
                \end{tabular}  
            \\ 
        \end{tabular}
    \end{tabular}%
    }
    \caption{Example response outputs for a dialogue context (DC) from each model, including commonsense where applicable. CS-\texttt{E/I} denote ConvoSense-\texttt{E/I}. Example explanations provided by human evaluators for the preference for ConvoSense-\texttt{E} against all other models are also shown (EX), along with the \textit{predicted aspects} from the automatic identification procedure and their \textbf{corresponding mapped categories}.}
    \label{tab:output-examples}
    \vspace{-3ex}
\end{table*}

\end{document}